# ALHD: A Large-Scale and Multigenre Benchmark Dataset for Arabic LLM-Generated Text Detection

Ali Khairallah[a,*], Arkaitz Zubiaga[a]

[a]*School of Electronic Engineering and Computer Science, Queen Mary University of London, London, United Kingdom*

## Abstract

We introduce ALHD, the first large-scale comprehensive Arabic dataset explicitly designed to distinguish between human- and LLM-generated texts. ALHD spans three genres (news, social media, reviews), covering both MSA and dialectal Arabic, and contains over 400K balanced samples generated by three leading LLMs and originated from multiple human sources, which enables studying generalizability in Arabic LLM-genearted text detection. We provide rigorous preprocessing, rich annotations, and standardized balanced splits to support reproducibility. In addition, we present, analyze and discuss benchmark experiments using our new dataset, in turn identifying gaps and proposing future research directions. Benchmarking across traditional classifiers, BERT-based models, and LLMs (zero-shot and few-shot) demonstrates that fine-tuned BERT models achieve competitive performance, outperforming LLM-based models. Results are however not always consistent, as we observe challenges when generalizing across genres; indeed, models struggle to generalize when they need to deal with unseen patterns in cross-genre settings, and these challenges are particularly prominent when dealing with news articles, where LLM-generated texts resemble human texts in style, which opens up avenues for future research. ALHD establishes a foundation for research related to Arabic LLM-detection and mitigating risks of misinformation, academic dishonesty, and cyber threats.

*Keywords:*

---

*Corresponding author

*Email addresses:* `a.m.khairallah@qmul.ac.uk` (Ali Khairallah),
`a.zubiaga@qmul.ac.uk` (Arkaitz Zubiaga)

LLM detection, Arabic NLP, Benchmarking, ALHD, Cyberthreats, Dialectal Arabic, Human vs AI text classification

---

## 1. Introduction

The rapid adoption of Large Language Models (LLMs) has advanced into our daily life routine across many fields such as education, medicine, and industry, leading to new levels of productivity and analytical power [1, 2], primarily thanks to the ease of generating high-quality content at scale. However, these advancements have introduced substantial cyber threats like misinformation, phishing, and academic dishonesty [3].

Advancements with LLMs have provided attackers more powerful tools for automating, personalizing, and persuasive text-based cyber threats, which can be maliciously used to create fake news, spams, phishing, or social engineering attacks [4]. This introduces new gaps that require further research to avoid misuse, not least the detection of unacknowledged uses of such LLMs for content generation [5].

The risks arising from LLM-based content generation are particularly crucial for linguistically sensitive scripts like Arabic, where even minor alterations in text can cause significant changes in meaning, especially in legal, religious, official documents, and even daily communication [6, 7].

Recent research has explored a range of approaches for LLM-generated text detection, including classical machine learning models such as Support Vector Machines (SVM) and Random Forest (RF), and more recent transformer-based architectures like BERT and its Arabic-specific variants [8, 9, 10, 11, 12]. Watermarking techniques, either injected or detected as passive semantic statistics have also shown promise in some contexts [13, 14, 15]. However, robust and non-invasive detection of LLM-generated Arabic text with strong cross-genre and cross-dialect generalization remains largely unexplored.

Considering the growing capabilities of LLMs, current detection approaches often fall short in robustness and generalizability, particularly as LLM outputs became increasingly coherent, natural, and human-like [16, 17, 7]. This creates pressing concerns about adversarial uses of LLMs. A significant gap hindering research in this direction is the lack of a large-scale, publicly available Arabic dataset spanning multiple genres and dialects annotated for both human- and machine-generated content representing a significant gap [18, 19, 20]. Addressing these challenges is essential for the advancement of a trustworthy Arabic NLP and effective cyber threat mitigation.

This work introduces the **Arabic LLM and Human Dataset (ALHD)**, a large-scale, multi-genres, and multi-dialect corpus specifically designed to distinguish between human- and LLM-generated texts with generalizability as the key objective. The ALHD dataset, constructed from six diverse Arabic sources from three genres and augmented with outputs from multiple leading LLMs, ALHD fills a critical gap in available resources and enables systematic benchmarking of detection methods. We present baseline experimental results on the dataset using a range of classical and transformer-based models, establishing new benchmarks for this crucial research problem. We make the ALHD dataset and benchmarking scripts publicly available to support ongoing research in Arabic AI-generated content detection and cyber threat mitigation. In addition, this work presents a benchmarking of state-of-the-art machine learning and transformer-based models on the constructed dataset. We provide a sophisticated analysis of the results, highlighting model strengths, weaknesses, and genre-specific challenges, as well as lessons learned and limitations that point to future research needs. To ensure reproducibility, we detail the steps taken in data set construction, preprocessing, and experimental design and setup.

The contributions of this work are threefold:

- we construct ALHD, the first large-scale comprehensive and balanced dataset of human- and LLM-generated Arabic texts spanning multiple sources, genres, and dialects.

- we benchmark a wide range of machine learning, BERT-based, and LLMs under multigenre and cross-genre setups.

- we provide a detailed analysis of the results. Outlining key lessons, limitations, and a roadmap for future research directions on Arabic LLM detection.

Our ALHD dataset provides a novel and unique resource enabling to further research in generalizable LLM-generated text detection in the Arabic language. Through our experimentation, we find that fine-tuned BERT-based models achieve the highest accuracy (often above 90%), traditional machine learning approaches provide a competitive but weaker baseline, and the prompting-based LLMs perform the worst, struggling to generalize across genres and dialects. Results show that larger training data leads to classification performance, yet cross-genre robustness remains a key challenge. We observe an increased challenge when the cross-genre detection is applied on news data, a genre in which human- and LLM-generated texts may be more similar to each other and therefore may need exploration of more methods to achieve competitive detection.

## 2. Background

### 2.1. Misuse of LLM-generated texts

Modern LLMs are increasingly capable of generating highly fluent human-like texts and adaptive to multiple dialects across genres [21]. While this progress unlocks many opportunities towards resolving daily life challenges, it also introduces risks in distinguishing between human- and machine-generated texts [22]. Undetected contents can cause serious cyber threats including misinformation, academic dishonesty, and even more aggressive consequences with phishing, smishing, and social engineering, where convincing texts are often used to manipulate individuals and organizations [23, 24]. For instance, the Anti-Phishing Working Group (APWG) have indicated in their phishing activity trends report that 932,923 phishing attacks were recorded worldwide in the third quarter of 2024, highlighting a significant increase in smishing by 22% during the same period. APWG also reported that social media platforms were the most targeted sector, representing 30.5% of all phishing attacks [24]. Furthermore, according to the FBI Internet Crime Report, losses due to cybercrimes in the United States have exceeded $12.5 billion in 2023 [25].

These examples illustrate how malicious manoeuvres can exploit machine-generated texts to scale deception. Hence, robust methods for detecting LLM-generated texts are urgently needed particularly with linguistically sensitive scripts such as Arabic, where formal, legal, and religious texts require extra reliability.

### 2.2. Arabic linguistic challenges

Arabic is spoken by over 1.6 billion people worldwide [26], and known by its various linguistic characteristics, such as the right-to-left text direction, absence of capital letters, variations in connectivity of different letters. Arabic employs diacritical symbols called *Harakat* to assist with pronunciation and comprehension of words, In many instances, two identical words may convey different meanings based on their diacritics, for example: جَمَل (Jamal: Camel) or جُمَل (Jumal: Paragraphs). In addition, a special character used in Arabic is the *Kashida* or *Tatweel*, which is represented by the presence of an elongation character (ـ) and which increases the length; however, it does not alter the meaning of the word or sentence, as for example, both سماعة and سماعة mean 'speaker' (Samā'a) [27, 26].

The use of Arabic scripts extends beyond daily and social interactions and professional contexts, and is also related to religious matters, including the Holy Quran and the Prophet's hadiths. In addition, legal and historical documents are considered sensitive and fragile in which any minor alteration in these texts may lead to significant changes in meaning [28, 27, 29]. This explicitly highlights the necessity of having linguistically sensitive LLM-detection methods, particularly for Arabic. Despite the challenges, current authentication approaches do not explicitly preserve semantics and do not generalize to different detection tasks in Arabic scripts, which highlights a notable gap in existing approaches.

## 3. Related Work

### 3.1. Datasets resources for Arabic Human and LLM text Detection

Detection of LLM-generated texts has gained significant attention, especially with the rapid progress in generative AI. While English and some other languages benefit from comprehensive resources, there is a notable lack of large-scale, multigenre Arabic datasets labeled for both human- and LLM-generated content, which poses a significant limitation to study generalizability of detection methods. Prior to ALHD, Arabic NLP lacked dedicated resources for human and LLM-generated texts in Arabic. In what follows, we discuss available datasets for Arabic LLM-generated text detection and their characteristics:

***AIRABIC.*** A recent effort by Alshammari and El-Sayed [30] introduced AIRABIC, a benchmark dataset of 1,000 Arabic samples, equally split between 500 human-written texts drawn from 41 diverse sources and 500 texts generated by GPT-3.5-turbo. The dataset provided both diacritized and non-diacritized forms of texts which was used to evaluate the classifiers under different conditions. While AIRABIC provides an evaluation using state-of-the-art classifiers such as GPTZero and OpenAI's text classifier, its limited size and use of single-generator highlight the necessity for a large, multi-genre Arabic dataset to support more robust benchmarking and cross-genre generalization.

**Darwish et al..** An earlier initiative by Darwish et al. [31] targeted social media, where the authors created a binary dataset consisting of 4,196 human-written Arabic tweets and 3,512 auto-generated tweets produced using GPT2-Small-Arabic. This dataset demonstrated that transformer classifiers can reach high accuracy (above 95%) in distinguishing human vs. GPT-2 texts, providing the first proof-of-concept in Arabic LLM detection.

**Okaz Fake News Dataset..** Another contribution is the Okaz Fake News Dataset [32], developed from the Saudi Okaz news outlet. It includes real human-written news articles combined with both GPT-generated fake articles and human-fabricated fake news. The dataset is structured as a three-class corpus (real, human-fake, and LLM-fake) and comprises more than 1,500 samples across genres such as politics, economics, health, sports. A GPT model was used as the text generator, providing an early but important attempt at integrating AI-generated Arabic news content into a detection resource.

**The Arabic AI Fingerprint Dataset..** This dataset [17] takes a different direction by focusing on stylometric signals of machine and human texts. It consists of 3,000 human-authored academic abstracts and 3,500 human-written social media reviews, each paired with machine-generated content produced by multiple LLMs, including ALLaM, Jais, LLaMA, and GPT-4. This design provides parallel human and machine corpora across two genres, enabling detailed stylometric analysis and highlighting that machine-generated texts are often exhibits detectable. Still, cross-genre evaluation remains an open challenge.

**Other datasets..** Other available Arabic corpora focus on specified genres like news, social media, and user reviews but are not explicitly labeled for distinguishing human-written from LLM-generated texts. These include AFND [20], ANAD [33], SANAD [34], BRAD [35], HARD [36], and MASC [37]. While valuable for a wide range of NLP tasks, they typically contain only human text, lack cross-genre diversity, and do not provide LLM-generated annotations. In our work, we leverage and build upon some of these datasets, expanding and adapting them for our aims to tackle generalizable LLM-generated text detection in Arabic.

This absence presents a critical barrier to developing and evaluating robust LLM-detection systems for Arabic [10, 11]. Table 1 shows a comparison of existing datasets and our new ALHD dataset, highlighting the higher order of magnitude in size and the higher diversity in terms of genres and language variants. ALHD addresses these gaps by introducing the first large-scale, balanced dataset exceeding 400K samples, explicitly annotated for human- and LLM-generated texts, spanning three genres (news, social media, reviews) and multiple dialects including Modern Standard Arabic (MSA) and Dialectal Arabic (DA), with texts generated using three leading LLMs in addition to human texts. This ensures both the diversity and the scale required to support systematic benchmarking and cross-genre evaluation

| Dataset | Size | Generators | Labels | Genres | Language variants |
|---|---|---|---|---|---|
| AIRABIC [30] | 1,000 samples (500 human, 500 machine) | GPT-3.5-Turbo | Human vs. LLM | Books and news articles | MSA + CA |
| gpt2-small-arabic [31] | 7,708 samples (4,196 human, 3,512 machine) | GPT2-Small-Arabic | Human vs. LLM | Social media (tweets) | DA |
| Okaz dataset [32] | 1,500 samples (real, human-fake, LLM-fake) | GPT (unspecified version) | Real vs. Human-fake vs. LLM-fake | News (politics, economics, health, sports) | MSA |
| The Arabic AI Fingerprint [17] | 6,500 human texts (3,000 academic, 3,500 reviews) + paired LLM outputs | ALLaM, Jais, LLaMA, GPT-4 | Human vs. LLM | Academic abstracts, reviews | MSA |
| **ALHD (ours)** | 405,456 samples (balanced) | GPT-3.5-Turbo, Gemini-2.5-Flash, Command-R | Human vs. LLM | News, social media, reviews (multigenre) | MSA + DA |

Table 1: Comparison of publicly available Arabic datasets containing both human- and LLM-generated texts.

## 3.2. LLM-generated Text Detection Methods

Early efforts on detection relied on classical machine learning approaches, such as Support Vector Machines (SVM), Logistic Regression (LR), and Random Forests (RF) [38, 39, 22]. While these methods achieved moderate success in distinguishing between human and machine-generated content in English, their performance deteriorates with increasingly advanced LLM outputs. Transformer-based models, particularly BERT, AraBERT, MAR-BERT, and ARBERT, have become the new standard for text classification tasks in Arabic, demonstrating superior performance due to their ability to capture complex semantic and morphological features [9, 11, 40, 18, 41, 8, 42, 43].

Watermarking-based methods have emerged as a complementary detection strategy, either by injecting invisible patterns (e.g., kashida, diacritics) or by leveraging passive statistical features in the generated text [26, 14, 15, 44, 45, 46]. While effective in certain scenarios, these techniques can suffer from limitations such as reduced fluency, ease of removal, or dependence on linguistic features that are not always present in Arabic scripts.

### 3.3. Benchmarks, Evaluation, and Identified Gaps

Large-scale benchmarking initiatives such as LAraBench [11] evaluate Arabic LLMs and BERT-family models across a range of tasks, revealing that transformer models, particularly BERT-based models, outperform classical ML baselines. However, these benchmarks primarily focus on standard NLP tasks and do not specifically address human vs. LLM-generated content detection. Boutadjine et al. [10] and Alshammari et al. [8] provided baseline results for AI-generated text detection in Arabic but relied on limited or synthetic datasets, and did not consider cross-genre experimental settings or adversarial evaluation. The lack of publicly available, large-scale, multigenre, and multi-LLM annotated Arabic corpora represents a major challenge for progress in this area, which our work addresses by introducing and experimenting with ALHD.

### 3.4. Summary

Despite significant advances, robust and generalizable detection of LLM-generated texts in Arabic remains an open research problem. Limitations in available datasets, the genre-specific nature of existing benchmarks, and the absence of adversarial or cross-genre evaluation underscore the need for new resources and systematic benchmarking. The present work addresses these gaps by introducing a comprehensive, multigenre Arabic dataset with human- and LLM-generated text labels and providing baseline results for multiple detection approaches.

## 4. Dataset Construction

We next describe the collection process to develop ALHD. First, we relied on existing datasets of human-written Arabic texts (Section 4.1), then expanded it to include diverse LLM-generated texts (Section 4.2), and finally cleaned the dataset (Section 4.3).

### 4.1. Data Collection

**Data sources for Human-generated texts in Arabic**. To obtain the subset pertaining to human-written texts in ALHD, we first identify six publicly available Arabic data sources, covering 3 different genres: news, social media, and reviews. This strategy ensures having multiple dialects, a large dataset size, and a diverse range of text lengths to support robust and generalizable evaluation. Details about these six datasets are described next and summarized in Table 2.

- **News articles #1 - SANAD** [34][1] - Over 190,000 news articles across 7 categories: culture, finance, medical, politics, religion, sports, tech. (Alarabiya lacks religion category) from Al-Khaleej, Al-Arabiya, and Akhbarona, all written in MSA.

- **News articles #2 - ANAD** [33][2] - Over 500,000 news articles from 12 Arabic news websites, all written in MSA.

- **Social media #1 - MDAT** [19][3] - Over 50,000 multi-dialect Arabic tweets covering multiple genres including politics, health, social, sports, economics, and other. This dataset contains texts written in DA, specifically Algerian, Egyptian, Lebanese, Moroccan and Tunisian dialects.

- **Social media #2 - MASC** [37][4] - 224,514,954 Arabic tweets written in both MSA) and DA.

---

[1]https://data.mendeley.com/datasets/57zpx667y9/2

[2]https://github.com/alaybaa/ANAD-Arabic-News-Article-Dataset/tree/main

[3]https://www.kaggle.com/datasets/um6popendata/sentiment-analysis-for-sm-posts-in-arabic-diale resource=download

[4]https://huggingface.co/datasets/pain/Arabic-Tweets

- **Reviews #1 - HARD** [36][5] - 400,000+ hotel reviews in Arabic, retrieved from booking.com and written in both MSA and DA.

- **Reviews #2 - BRAD** [35][6] - 510,600 book reviews in Arabic, retrieved from the Goodreads website and written in both MSA and DA.

| # | Name | Size | Genre | Dialect |
|---|------|------|-------|---------|
| 1 | SANAD | +190,000 | News | MSA |
| 2 | ANAD | +500,000 | News | MSA |
| 3 | MDAT | +50,000 | Social Media | DA |
| 4 | MASC | 224,514,954 | Social Media | MSA & DA |
| 5 | HARD | +400,000 | Reviews | MSA & DA |
| 6 | BRAD | 510,600 | Reviews | MSA & DA |

Table 2: Summary of Arabic Text Datasets

*Note: All mentioned datasets are publicly available. MSA stands for Modern Standard Arabic, DA stands for Dialectal Arabic.*

***Data cleansing***. With each of the six data sources above, we perform a data cleansing process, which includes removal of duplicated texts, dropping empty or null fields, and discarding texts with fewer than 4 tokens (using AraBERT tokenizer [9]). This process ensures consistency and fair sampling across datasets.

***Sampling, merging, and reshaping***. We compose a dataset of 100,000 human- and 300,000 LLM-generated texts, evenly distributed across the six available sources; to achieve this, we define a sample size of about 16,667 samples per source. However, to reach a total of 100,000 samples and to account for issues such as potential LLM generation anomalies, e.g. empty strings, duplication, and failed requests, we increased the sample size to 17,000 per source (total 102,000). Subsequently, all the sampled data is merged in one main set to form the first part (human-generated texts) of ALHD.

---

[5] https://github.com/elnagara/HARD-Arabic-Dataset
[6] https://github.com/elnagara/BRAD-Arabic-Dataset

### 4.2. Process for Generating LLM Texts

To ensure comprehensive and meaningful benchmarking in detection of human and LLM generated texts, we selected three distinct and widely used LLMs: GPT[7] (OpenAI), Gemini[8] (Google), and Command[9] (Cohere), these models are among the most adopted, technically advanced, used by individuals or enterprises. OpenAI GPT-3.5-turbo is recognized for its generative text capabilities and has set of benchmarks in NLP tasks [47], Google Gemini-2.5-flash represents one of the best performed models in adaptive thinking [48], and Cohere Command-R, a multilingual highly performing retrieval-augmented generation (RAG) [49] as shown in Table 3.

| # | Model | Input Type | Output Type | Context Window | Max Output Tokens |
|---|---|---|---|---|---|
| 1 | GPT-3.5 Turbo | Text | Text | 16,385 tokens | 4,096 tokens |
| 2 | Gemini 2.5 Flash | Text, images, video, audio | Text | 1,048,576 tokens | 65,536 tokens |
| 3 | Command-R | Text | Text | 128,000 tokens | 4,000 tokens |

Table 3: Comparison of LLMs used in this work. All models support Arabic input and output.

To standardize the generation process, we employed a unified prompting strategy across all LLMs. For every model, the following instruction (in Arabic, with the translation provided right after) was used as the main user prompt:

---

**User prompt in Arabic:**

"اكتب نصاً عربياً جديداً عن نفس موضوع النص التالي وبنفس اللهجة المستخدمة وبطول مقارب للنص الأصلي دون إعادة صياغة أو تلخيص النص الأصلي."

---

> **User prompt in English:**
>    "Write a new Arabic text on the same topic as the following text, in the same dialect used, and with a length similar to the original text, without paraphrasing or summarizing the original text."

To reflect regular user interactions with LLMs, no additional generation parameters were changed (e.g. temperature, top-p, max-tokens), and we used the API's default settings for each model. For GPT-3.5-turbo and Command-R, the prompt was composed using both system and user messages, as these models support and are sensitive to system-level instructions.

> **System prompt in Arabic:**
> "أنت مساعد لغوي محترف ومبدع. مهمتك كتابة نصوص عربية جديدة وطبيعية بالكامل، وعدم الاعتماد على إعادة الصياغة أو النسخ من النص الأصلي، بل كتابة نص أصيل كما لو كان مكتوبًا من كاتب آخر في نفس السياق واللهجة وطول النص"

> **System prompt in English:**
>    "You are a professional and creative linguistic assistant. Your task is to write completely new and natural Arabic texts, without relying on rephrasing or copying from the original text. Instead, you should write an original text as if it were written by another author in the same context, dialect, and length as the original."

For Gemini-2.5-flash, only a user prompt was used, as the API does not support a system prompt. This approach ensures that the results represent standard out-of-the-box LLM behavior and are directly comparable to common usage patterns in real life.

### 4.3. Clean, merge, and balance final dataset

After generating LLM content, a sanity check is performed including checking shape, assess missing/null values, empty strings, and duplicated texts. We then perform a cleaning process to remove any null values, empty strings, in-dataset and cross-dataset deduplication. The resulting data is then finally balanced to obtain the final dataset as summarized in Table 4.

***Data balancing***. ALHD dataset has ratio of 1:3 (Human to LLM), so to generate fully balanced dataset, we have to make sure it is balanced across different aspects including label, source, and generator. This approach ensures no bias happens in benchmarking. The first step is to calculate minimum document IDs per source using the following formula:

Let $S$ be the set of all sources, and $D_s$ denote the set of eligible document IDs for source $s$. The number of document IDs to select per source is computed as:

$$n_{\text{per\_source}} = \min_{s \in S} |D_s|$$

where $|D_s|$ is the number of eligible document IDs for source $s$.

After calculating number of documents per source, we sample this number once for human-written texts, and random sample from its equivalent LLM-generated texts with same document ID.

In the final dataset, each entry in the dataset is structured to contain the following fields, with the description of each field and its possible values provided in Table 5:

- **text:** Full Arabic text sample.

- **label:** Indicates whether the text is human-generated, indicated by 0, or LLM-generated, indicated by 1.

- **source:** The original data source or platform which the text was obtained (for the human source) or the human source used for LLM generation. Example values: SANAD, ANAD, MASC, MDAT, HARD, and BRAD, for more information, see Table 2.

- **generator:** The name of the model used to generate LLM text with possible values: GPT-3.5-turbo, Gemini-2.5-flash, Command-R, and Human.

- category: The main subject area of a text with possible values: Twitter, Goodreads, Booking.com, alwatan.com.sa, etc.

- **subcategory:** The specific subgenre within the main category with possible values: Economy, Local, Sport, Tweet, etc.

- **token count:** Indicating the number of tokens (words or wubwords as defined by the tokenizer) in the text sample, to generate this we use the AraBERT tokenizer to tokenize each sample then the number of tokens is stored as token count.

- **document ID:** A unique identifier that links the text sample to its original document or record. Each human-written and its corresponding LLM-generated text will have the exact unique number, which can help to identify the source text.

| Source | COMMAND-R | GEMINI-2.5-F | GPT-3.5-T | HUMAN |
|--------|-----------|--------------|-----------|-------|
| ANAD | 16,894 | 16,894 | 16,894 | 16,894 |
| BRAD | 16,894 | 16,894 | 16,894 | 16,894 |
| HARD | 16,894 | 16,894 | 16,894 | 16,894 |
| MASC | 16,894 | 16,894 | 16,894 | 16,894 |
| MDAT | 16,894 | 16,894 | 16,894 | 16,894 |
| SANAD | 16,894 | 16,894 | 16,894 | 16,894 |
| **Total** | **101,364** | **101,364** | **101,364** | **101,364** |
| Grand Total: 405,456 | | | | |

Table 4: Number of samples per source and generator in the final balanced ALHD dataset. Each source contains an equal number of samples from every generator (human and LLMs).

## 5. Dataset Analysis

The final ALHD dataset is composed of 405,456 texts derived from six sources across three genres, with a diverse set of genres spanning 18 categories and 19 subcategories, and produced by 4 generators (3 LLMs + 1 human).

Each sample is identified by a document_id (101,346 unique samples) and a generator (4) pair, making up the total above. By definition, every document_id has one human reference and multiple LLM variants, which yielded 25% Human (n= 101,346) vs 75% LLM (n= 304,092) (1:3). Token counts (calculated by AraBERT tokenizer), are heavy-tailed and vary substantially by source; extreme values include BRAD with a maximum of 88,469 tokens, with the detailed token ratios shown in Table 6, and summarized in Figure 1.

The final statistics of the dataset are shown in Table 7.

Figure 2 shows lexical fingerprints per generator. Columns are Human, GPT-3.5-T, GEMINI-2.5-F, and COMMAND-R, rows are ranked (1-10). Each cell points the top token with its corpus frequency (k), the background color encodes $\log_{10}$(count). We removed Arabic stopwords without stemming or lemmatization and aggregation of counts over all texts per generator. The pattern shows a consistency in generator direction: human-generated texts have highly frequent formal tokens in MSA and numerals such as خلال (during), العام (the year), and مليون (million). In contrast, GPT-3.5-T leans towards connective, adverb, and modality words like بشكل (in a way), يمكن (may/can), دون (without), and مثل (such as). GEMINI-2.5-F and COMMAND-R introduce dialectal and conversational markers at high ranks including اللي (that/ which), يا (Calling someone: O/hey), and فيها/فيه (there is/ there are/ in it [m/f]). Notably, الله (God) remains frequent across all generators, reflecting ALHD's rich and realistic expression mix.

| Column | Description | Possible Values |
|---|---|---|
| `text` | Arabic full sample text, either human-written or generated by an LLM | Any natural-language text in Arabic |
| `label` | Distinguishes human vs LLM-generated text | 0 = human-written<br>1 = LLM-generated |
| `generator` | Generator identity | `GPT-3.5-T`, `GEMINI-2.5-F`, `COMMAND-R`, `HUMAN` |
| `source` | Dataset source/corpus | `SANAD`, `ANAD`, `MDAT`, `MASC`, `HARD`, `BRAD` |
| `category` | Source-specific main topic/category | Defined by original source, e.g., Twitter, Goodreads, bbc.com:arabic, etc. |
| `subcategory` | Source-specific subcategory | Defined by original source, e.g., Art, Health, Sport, etc. |
| `token_count` | Number of tokens in text (using AraBERT tokenizer) | Positive integer |
| `document_id` | Unique identifier for each original document, each appears exactly 4 times (1 human + 3 LLMs) in the dataset | Positive integer |

Table 5: Description of columns and value domains in the final ALHD dataset.

| Source | $n$ | Min | Q1 | Median | Q3 | Max |
|---|---|---|---|---|---|---|
| SANAD | 67,576 | 5 | 254 | 356 | 499 | 6,683 |
| ANAD | 67,576 | 7 | 250 | 347 | 483 | 6,522 |
| MDAT | 67,576 | 1 | 17 | 45 | 126 | 4,551 |
| MASC | 67,576 | 2 | 22 | 55 | 125 | 4,076 |
| BRAD | 67,576 | 1 | 57 | 131 | 260 | 88,469 |
| HARD | 67,576 | 2 | 28 | 61 | 129 | 4,063 |

Table 6: Per-source token count statistics. IQR is shown via Q1 and Q3.

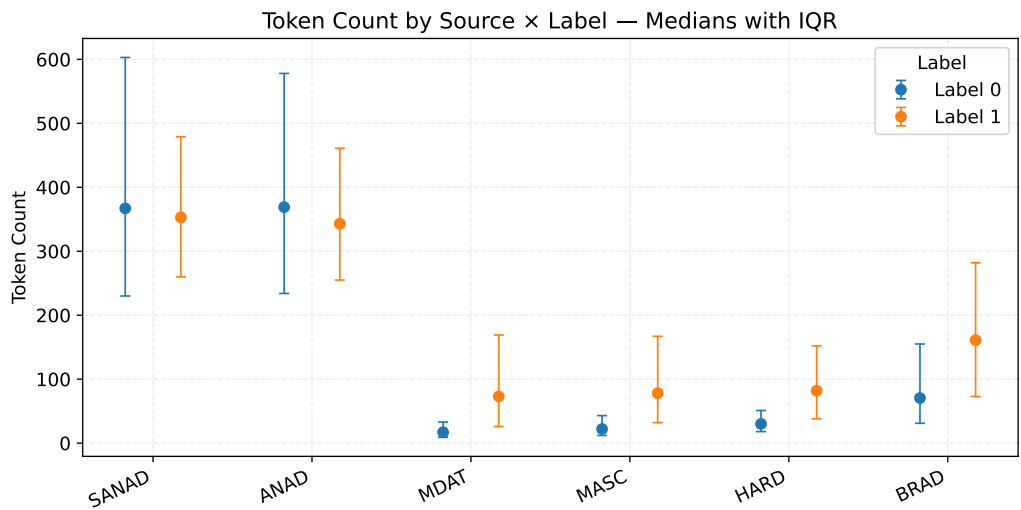

Figure 1: Token counts by source and label, summarized with medians and interquartile ranges. This highlights distributional differences between Human (label 0) and LLM (label 1) across sources.

| Field | Unique values |
|:---:|:---:|
| text | 405,445 |
| label | 2 |
| generator | 4 |
| category | 18 |
| subcategory | 19 |
| source | 6 |
| token_count | 2,443 |
| document_id | 101,364 |

Table 7: Final statistics of the ALHD dataset.

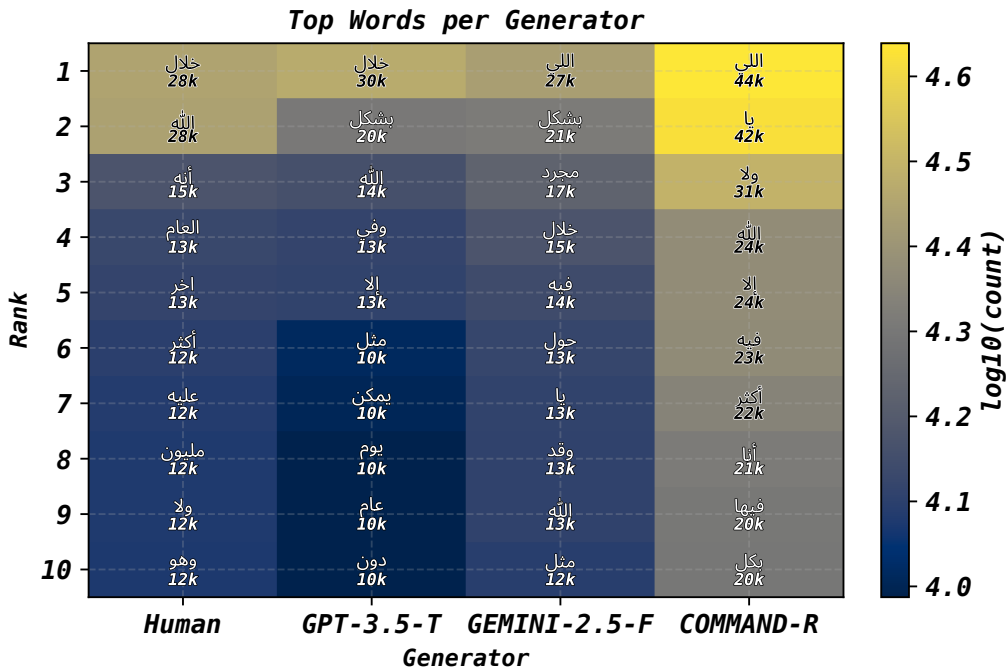

Figure 2: Top-10 content words per generator (columns: Human, GPT-3.5-T, GEMINI-2.5-F, COMMAND-R). Each cell shows the highest-ranked token (Arabic printed as-is) with its corpus frequency (k), the background encodes $\log_{10}$(count).

## 6. Experimental Setup

To ensure reproducibility and consistency across all experiments, fixed random seeds and standardized environment configurations were applied throughout the workflow. Consistent sanity checks were performed before and after each major processing step to guarantee no data leakage, imbalance (in label, source, or generator), or inconsistency between datasets. Furthermore, advanced logging was implemented to capture all experiment configurations, key actions, and results, enabling transparent tracking and fair benchmarking.

### 6.1. Construction of 10% balanced subset

Alongside the full balanced dataset, we provide a 10% balanced subset. This smaller split enables fast experimentation, hyperparameter tuning, and benchmarking under resource constraints, which would be otherwise infeasible with a full-sized dataset. In addition, it enables systematic analysis of model responsiveness to dataset size by directly comparing the results on the 10% and full sets. This dual-sized split allows us not only to evaluate the impact of dataset volume on performance, but also to analyze whether the gains are a result primarily of the increased training data or of model capacity.

Using generated fully balanced dataset, we construct the 10% subset by doing 3 steps:

- Stack: All LLM-generated rows are stacked next to its respective human-generated rows using document ID as key.

- Sample: Sample 10% balanced source subset.

- Unstack: return all LLM-generated back as rows.

### 6.2. Data splitting: train, validation, test

The ALHD balanced datasets was subjected to a strict splitting pipeline to ensure balanced and high-quality data for all subsequent experiments. Below, we detail each step and the key formula or algorithms employed. Balanced splitting is performed to assess full dataset benchmarking and generalizability by using held out source(s) as test sets. For standard splits we use ratio of 70/15/15 train/validation/test partitions. For cross-source evaluation, we use ratio of 80/20 train/validation split on non-test sources, and reserve all held-out source(s) samples as the test set as shown below.

**Full split (no held-out sources)**

$$n_{\text{train}} = \left\lfloor \frac{N \times r_{\text{train}}}{n_{\text{sources}}} \right\rfloor$$

$$n_{\text{val}} = \left\lfloor \frac{N \times r_{\text{val}}}{n_{\text{sources}}} \right\rfloor$$

$$n_{\text{test}} = \frac{N}{n_{\text{sources}}} - n_{\text{train}} - n_{\text{val}}$$

**Held-out split (one or more test sources)**

$$n_{\text{train}} = \left\lfloor \frac{N_R \times r_{\text{train}}}{n_{\text{sources}} - h} \right\rfloor$$

$$n_{\text{val}} = \left\lfloor \frac{N_R \times r_{\text{val}}}{n_{\text{sources}} - h} \right\rfloor$$

$$n_{\text{test}} = N_H$$

**Where:**

- $N$ = Total number of samples in the dataset

- $n_{\text{sources}}$ = Number of unique sources in the dataset

- $h$ = Number of held-out sources

- $N_H$ = Total number of samples in the held-out source(s)

- $N_R$ = Total number of samples in the remaining sources ($N_R = N - N_H$)

- $r_{\text{train}}$ = Proportion assigned to training set (e.g., 0.7 or 0.8)

- $r_{\text{val}}$ = Proportion assigned to validation set (e.g., 0.15 or 0.2)

- $n_{train}$ = Number of training samples per non-held-out source

- $n_{val}$ = Number of validation samples per non-held-out source

- $n_{test}$ = Number of test samples (all held-out samples)

This step will generate a total of 10 sets of data, each with train, validation, and test split, and available in two portions: 10% and full sized dataset split. Generated sets include using all sources and held-out sources as following:

- Single-source: isolating 1 source as held-out in the test set, this is crucial to evaluate unseen data in train and validation. The test set will contain all held-out source samples (no capping), for more information please refer to Table 2.

- Multi-related sources: isolating two related sources (e.g. isolating news data means using SANAD and ANAD as held-out), this ensures the model is exposed to unseen and unrelated data. The test set will contain all held-out sources samples (no capping), for more information please refer to Table 2.

### 6.3. Implementation details

All experiments were implemented in Python using a modular and reproducible pipeline. The following summarizes the main aspects of the implementation:

***Experiment configuration.*** Experiments were parametrized with EXPERIMENT_TYPE (traditional, BERT-based, or LLM-based), TEST_SOURCES (held-out sources used exclusively in test sets), and DATA_SIZE (full balanced dataset or 10% subset). All configuration variables including file paths and log directory were programmatically constructed to ensure unique, timestamped logging in each run.

***Logging and traceability.*** Advanced timestamped logging was employed, capturing all critical configurations, library versions, and runtime feedback. Custom exception and warning handles were also captured, supporting full traceability and reproducibility.

**Environment, libraries, and Reproducibility**. Random seeds were fixed at the beginning and used at every stage to guarantee deterministic results. All library versions were logged at the start of each experiment, including Python, pandas, numpy, PyTorch, etc. Experiments were run using Python 3.x on a Linux-based environment with both CPU and GPU support.

**Data handling and integrity**. Training, validation, and test splits were loaded from standardized dynamic approach using experiment parameters. Sanity checks were implemented after each main step to ensure no document ID leakage between train, validation, and test splits, any detected leakage results a fatal error and is logged for reference.

**Computational resources**. This research utilized Queen Mary's Apocrita high-performance computing (HPC) facility, supported by QMUL Research-IT [50]

*6.4. Evaluation metrics*

Model performance was evaluated using several standard classification metrics. The primary metrics reported are accuracy, macro F1 score, and ROC-AUC, all computed on the test set. Accuracy reflects the proportion of correct predictions, while the macro F1 score forms the arithmetic mean of F1 scores for both classes ( human-written and LLM-generated text), thus providing a balanced assessment in presence of any class imbalance. The ROC-AUC (Receiver Operating Characteristic - Area Under the Curve) measures the model's ability to distinguish between two classes.

In addition to these metrics, we provide full classification report (including precision, recall, f1 score, support across classes) and the confusion matrix for detailed error analysis.

# 7. Benchmarking Tasks

To benchmark models using ALHD dataset, we have prepared several approaches to assess model performance and robustness. Our experiments are configured to evaluate in-dataset and cross-dataset settings, providing insights into the true model performance and generalization ability to unseen sources of data. We experiment with three state-of-the-art families of models: traditional machine learning models (e.g. Random Forest), BERT-based models (e.g. AraBERT), and large language models (e.g. GPT). The selection of models in each family is subjected by factors such as support for Arabic (or multilingual), model size, and prevalence in related work.

*7.1. Traditional machine learning models*

**Feature extraction**. Each input text is vectorized using TfidfVectorizer with max_features=10,000, unigrams and bigrams (ngram_range=(1,2) ), and default tokenization. Vectorization has been made separately for train/validation/test sets to ensure no data leakage.

**Model selection**. A set of baseline models in text classification such as Logistic Regression, LinearSVC, ComplementNB (a Naive Bayes variant), LightGBM, and Random Forest were used for benchmarking [51, 52]. These models were chosen to provide a robust baseline comparison, covering different categories including Linear models (Logistic Regression, Linear SVC), Probabilistic models (ComplementNB), and Tree-based ensembles (LightGBM, Random Forest). These models were chosen to cover major classical machine learning families and ensures that the baseline experiments can be directly benchmarked with BERT- and LLM-based methods evaluation results.

**Hyperparameter tuning**. Each model was subjected to hyperparameter tuning to maximize performance. Automated grid search (GridSearchCV, 3-fold stratified Cross Validation) was applied to the following key hyperparameters:

- **LogisticRegression**: C: [0.1, 1.0, 2.0], solver: [liblinear, saga], max_iter: [500, 1000]

- **LinearSVC**: C: [0.1, 1.0, 2,0], max_iter: [3000, 5000]

- **ComplementNB**: alpha: [0.01, 0.1, 0.5, 1.0], norm: [True, False]

- **LightGBM**: learning_rate: [0.1], num_leaves: [31, 63], max_depth: [10]

- **RandomForest**: n_estimators: [200, 400], max_depth: [None, 20], min_samples_leaf: [1, 2]

These parameter ranges were chosen to achieve efficient and robust evaluation results. All models were loaded, trained, and evaluated on the same data splits, random seeds, and standard experimental setup to ensure fairness in comparison.

### 7.2. BERT-based models

State-of-the-art BERT-based model backbones were loaded for benchmarking, including Arabic-specific such as AraBERTv2 [9], AraBERTv2-Large [9], AraELECTRA [41], ARBERTv2 [40, 53], MARBERTv2 [40], Asafaya-BERT-base [54], and Asafaya-BERT-large [54], as well as multilingual models such as Google-mBERT [55], XLM-RoBERTa-base [56], and XLM-RoBERTa-large [56] as shown in Table 8. Each model's associated tokenizer is loaded from HuggingFace. For certain models like AraBERT, AraBERTv2-Large, and AraELECTRA, we included preprocessing prior to tokenizing as recommended in their documentation, rest of models text scripts used without any modification.

### 7.3. Large language models (LLMs)

We benchmarked transformer-based LLMs on Arabic LLM vs human text classification in two settings: zero-shot and few-shot. Models were selected based on available resources, efficiency, and readiness for the Arabic, including multilingual models that support Arabic. Table 9 shows the models benchmarked in our baseline experiments.

Models were prompted and configured to output either 0 (human) or 1 (LLM) to ensure deterministic results by instructing in prompt and forcing to generate maximum of 1 token to ensure strict digit-only outputs.

With LLMs, we test two different setups, zero-shot (where the model is given an instruction but no samples) and few-shot (where the model is shown an instruction alongside a small number of labeled samples). We describe these two setups in more detail next.

***Zero-shot setup***. In zero-shot, no training or validation data were introduced, a special prompt was constructed as follows:

---

**ZS prompt in Arabic:**

مهمتك تصنيف النص إذا كان بشرياً (0) أو من إنتاج نموذج لغوي ضخم LLM (1).

اكتب الرقم فقط بدون أي شرح.

النص: <Text>

التصنيف:

---

| Model | Parameters | Language(s) | Pretraining Data |
|-------|-----------|-------------|------------------|
| AraBERTv2-Base[1] | 136M | Arabic | 77GB of Arabic news, Wikipedia, OSIAN, and other resources |
| AraBERTv2-Large[2] | 371M | Arabic | 77GB of Arabic news, Wikipedia, OSIAN, and other resources |
| AraElectra[3] | 136M | Arabic | 77GB of Arabic news, Wikipedia, OSIAN, and other resources |
| ARBERTv2[4] | 163M | Arabic | 243GB of multi-sourced MSA and Arabic part mC4 corpus |
| MARBERTv2[5] | 163M | Arabic | 198GB of Arabic tweets |
| Asafaya-BERT-Base[6] | 111M | Arabic | ∼95GB of Wikipedia, OSCAR, and other Arabic sources |
| Asafaya-BERT-Large[7] | 338M | Arabic | ∼95GB of Wikipedia, OSCAR, and other Arabic sources |
| Google-mBERT[8] | 110M | 104 languages incl. Arabic | Wikipedia data |
| XLM-R-Base[9] | 270M | 100 languages incl. Arabic | 2.5TB of CC100 (CommonCrawl corpus) |
| XLM-R-Large[10] | 550M | 100 languages incl. Arabic | 2.5TB of CC100 (CommonCrawl corpus) |

Table 8: Specifications of BERT-based models used in ALHD benchmarking. All models were obtained from HuggingFace.

[1]`aubmindlab/bert-base-arabertv2`    [2]`aubmindlab/bert-large-arabertv2`
[3]`aubmindlab/araelectra-base-discriminator`
[4]`UBC-NLP/ARBERTv2`    [5]`UBC-NLP/MARBERTv2`    [6]`asafaya/bert-base-arabic`
[7]`asafaya/bert-large-arabic`    [8]`google-bert/bert-base-multilingual-cased`
[9]`FacebookAI/xlm-roberta-base`    [10]`FacebookAI/xlm-roberta-large`

| Model | Parameters | Language(s) |
|---|---|---|
| Qwen2.5-7B-Instruct[1] | ~7B | Multilingual (29 languages incl. Arabic) |
| JAIS-13B-Chat[2] | 13B | Arabic (MSA) and English |
| GPT-OSS-20B[3] | 20B | Multilingual (incl. Arabic) |
| ALLaM-7B-Instruct-preview[4] | 7B | Arabic and English (advanced in Arabic) |
| c4ai-command-r7b-arabic-02-2025[5] | ~7B | Arabic and English (Optimized for Arabic MSA) |
| Gemma-3-12B-IT[6] | 12B | Multilingual (140 languages) |

Table 9: Specifications of LLMs used in ALHD benchmarking. All models were obtained from Hugging Face.
[1]`Qwen/Qwen2.5-7B-Instruct` [2]`inceptionai/jais-13b-chat` [3]`openai/gpt-oss-20b` [4]`ALLaM-AI/ALLaM-7B-Instruct-preview` [5]`CohereLabs/c4ai-command-r7b-arabic-02-2025` [6]`google/gemma-3-12b-it`

---

**ZS prompt in English:**

Your task is to classify the text as Human (0) or produced by a Large Language Model (1).

Write only the digit, with no explanation

Text: <text>

Label:

---

***Few-shot setup***. FS examples were drawn only from the training set using predefined configuration parameter for number of examples, in our experiments we configured number of examples to 8 (4 human and 4 LLM). The prompt was constructed similar to zero-shot, with the addition of examples as follows:

---

**FS Prompt in Arabic:**

مهمتك تصنيف النص إذا كان بشرياً (0) أو من إنتاج نموذج لغوي ضخم LLM (1).

اكتب الرقم فقط بدون أي شرح.

أمثلة:

النص: <example_0_text>

التصنيف: 0

```
...
النص: <example_1_text>
التصنيف: 1
النص: <Text>
التصنيف:
```

---

**FS prompt in English:**

Your task is to classify the text as Human (0) or produced by a Large Language Model (1).

Write only the digit, with no explanation

Examples:

Text: <example_0_text>

Label: 0

...

Text: <example_1_text>

Label: 1

Text: <text>

Label:

---

## 8. Evaluation Scenarios

We set up the experiments with test scenarios that evaluate both multi-genre and cross-genre generalization. To do so, we define a total of 10 scenarios as follows:

- **All sources (multigenre) split (1 scenario):** models are trained and tested using standard 70/15/15 train/validation/test splits. This helps to assess the in-genre baseline where train and test distributions come from the same pool.

- **Single-source (within-genre) isolation (6 scenarios):** each of the six sources (SANAD, ANAD, MDAT, MASC, HARD, BRAD) is held out in turn as test set, while the remaining will be used for train/validation in 80/20 splits. These scenarios directly test cross-source robustness by exposing models to unseen sources while still keeping the same genre represented in training.

- **Two-source (cross-genre) isolation (3 scenarios):** for each genre (news, reviews, social media), both sources are held out together to form the test set (e.g., SANAD+ANAD for news). This provides a strict cross-genre evaluation, where models must generalize to an entire genre excluded from training. The remaining four sources are divided into 80/20 train/validation splits.

In addition, all experiments will be evaluated on dual-scaled datasets: 10% balanced (~20k) and full balanced (~200k) datasets. The smaller split is not only practical for fast prototyping, hyperparameter tuning, and benchmarking under limited resources, but also plays an essential role in our experimental design. Using both reduced and full datasets allows us to assess the effect of data volume on model performance. By comparing results across the two scales, we can distinguish whether performance gains are primarily the result of increased training data or from model capacity.

## 9. Results and Discussion

In this section, we present the results of benchmarking three categories of models: traditional machine learning classifiers, BERT-based transformer models, and large language models (LLMs) on the constructed ALHD dataset. 540 experiments were conducted, including all listed models, data sizes, and splits. The analysis covers overall performance trends, category-specific strengths and weaknesses, the effect of dataset size, and robustness across different evaluation splits. In addition, we identify the best and worst models, and perform error analysis.

### 9.1. Overall performance across models

As we can see in Figures 3 and 4, we observe clear performance differences across the three model categories. BERT-based architectures achieved the highest mean accuracy (0.902), followed by traditional models (0.804), while large language models (LLMs) without fine-tuning underperforming others (0.497). This ranking was consistent across other evaluation metrics, with BERTs achieving mean ROC-AUC above 0.98, traditional models averaging 0.91 and LLMs failing to exceed 0.47. These results prove that fine-tuned BERT models as state-of-the-art for this task, with traditional models providing strong baselines and LLMs proving worst overall

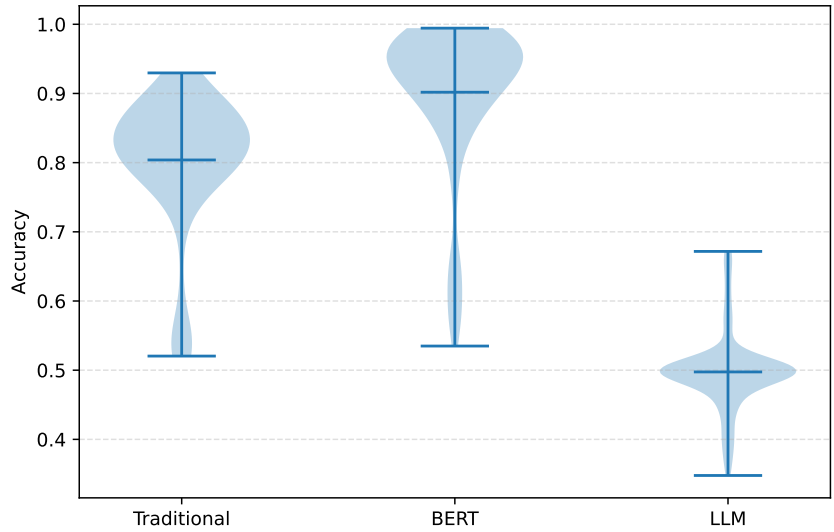

Figure 3: Overall accuracy distributions across model families. BERT-based models dominate, traditional models are strong baselines, and LLMs cluster near chance.

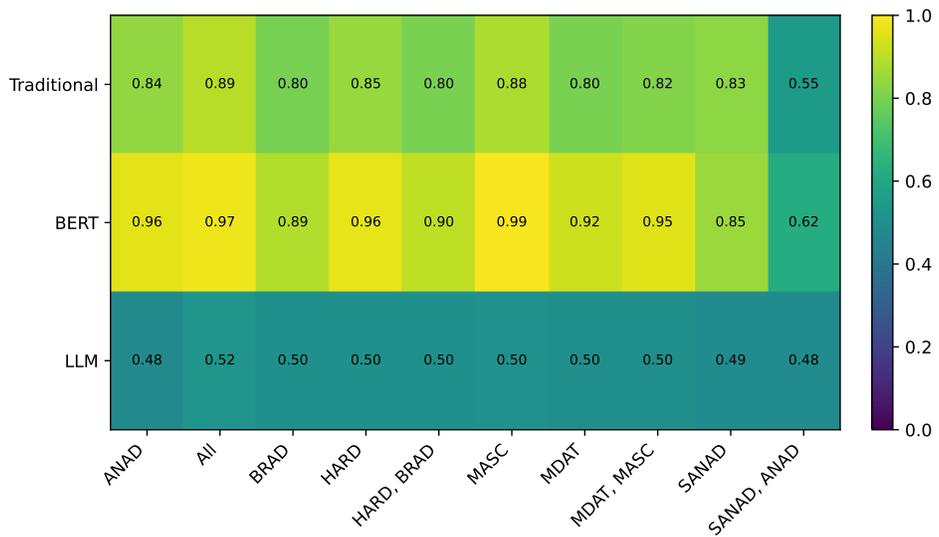

Figure 4: Mean accuracy by family across dataset splits (heatmap).

Looking at the results shown in Figure 5, traditional classifiers provided competitive performance, particularly Linear SVC and Logistic Regression, which achieved mean accuracies of 0.82 and 0.81 respectively. The best traditional model is a Linear SVC on the full dataset with all-source (multi-genre) split, achieving an accuracy of 0.93 and a ROC-AUC of 0.98. Logistic Regression and Random Forests come next, but still attained accuracies between 0.91 and 0.93. As traditional models provide fair performance, this can be considered appealing due to their computational efficiency and strong accuracy. However, performance degraded significantly on more challenging cross-genre splits (i.e. SANAD+ANAD), which contains large scripts derived from news sources written in MSA, where evaluation results range between 0.52-0.62, averaging below 0.55 accuracy due to the complexity of these texts.

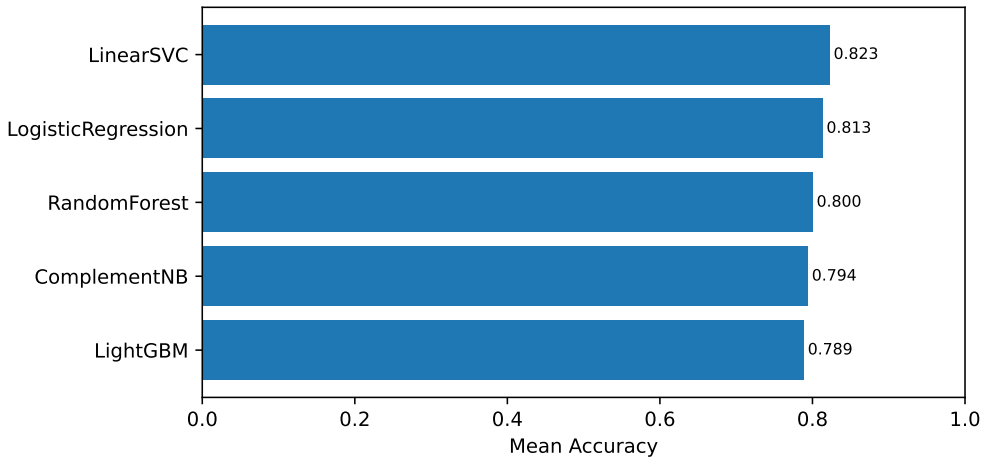

Figure 5: Traditional family: mean accuracy per model. LinearSVC and LogisticRegression lead, with RandomForest, ComplementNB, and LightGBM close behind.

### 9.3. BERT-based models

As we can see in Figure 6, BERT-family models dominated this benchmark, particularly with Arabic tailored variants such as AraBERTv2-Large, AraElectra, and AraBERTv2-Base consistently exceeded 0.91 mean accuracy, while multilingual models like XLM-R-Base achieved the single best run (0.994 accuracy). These results confirm that fine-tuned transformer encoders are highly effective in detecting morphological and semantical patterns between human- and LLM-generated texts. Genre effects were notable, for example in the social media dataset MASC, models achieved mean accuracy of 0.989. This performance declined on the news dataset SANAD to 0.76 with Google-mBERT, observing an even lower performance in cross-genre isolation (SANAD+ANAD) with an accuracy of 0.53. AraBERTv2-Large obtained the best generalizability and genre-adaptation in this case with 0.77 accuracy, which highlights the sensitivity of having a multiple genres in ALHD. Overall, BERTs combined high accuracy with balanced precision and recall, making them the most reliable category.

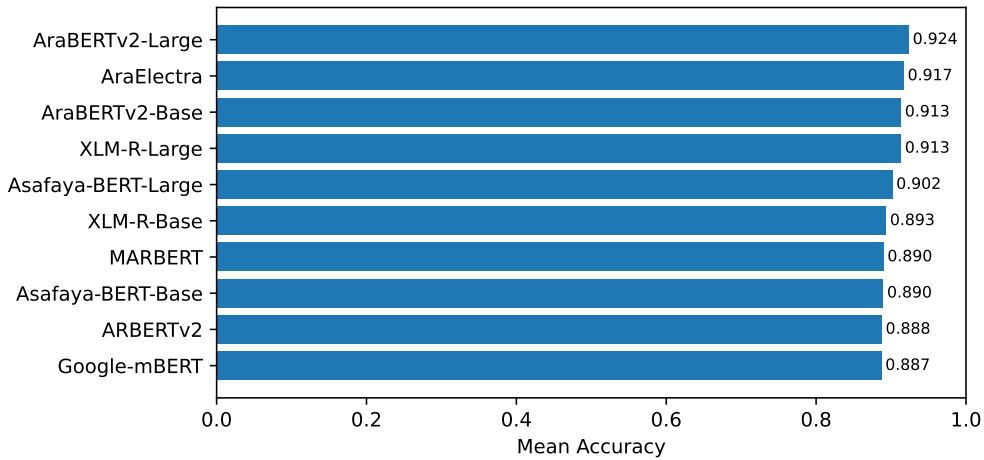

Figure 6: BERT family: mean accuracy per model. Arabic-pretrained variants (AraBERTv2, AraElectra) outperform multilingual alternatives.

### 9.4. Large language models

LLMs under zero-shot (ZS) and few-shot (FS) prompting consistently failed to match supervised approaches as shown in Figure 7. JAIS-13B-Chat was the strongest in ZS, averaging 0.622 accuracy, while most other LLMs achieved scores around 0.50. Gemma-3-9B in ZS frequently collapsed to accuracies below 0.40, which represents the weakest performance across all models. These findings indicate that prompting alone is insufficient and fine-tuning is necessary to unlock the potential of LLMs.

### 9.5. Overall results

Table 10 shows overall performance across families, consolidating all experiments as mean metrics $\pm$ 95% confidence interval (CI) to show variance, BERT-based encoder models dominate this task, with Arabic-tailored backbones including AraBERTv2-Large and AraElectra, achieving the highest mean accuracy of 0.92 and 0.91 respectively, and Macro F1 ranging between 0.91-0.92. Multilingual encoders (XLM-R) remain highly competitive, proving that fine-tuning helps to capture rich lexical and morphological features that separate human and LLM text in Arabic.

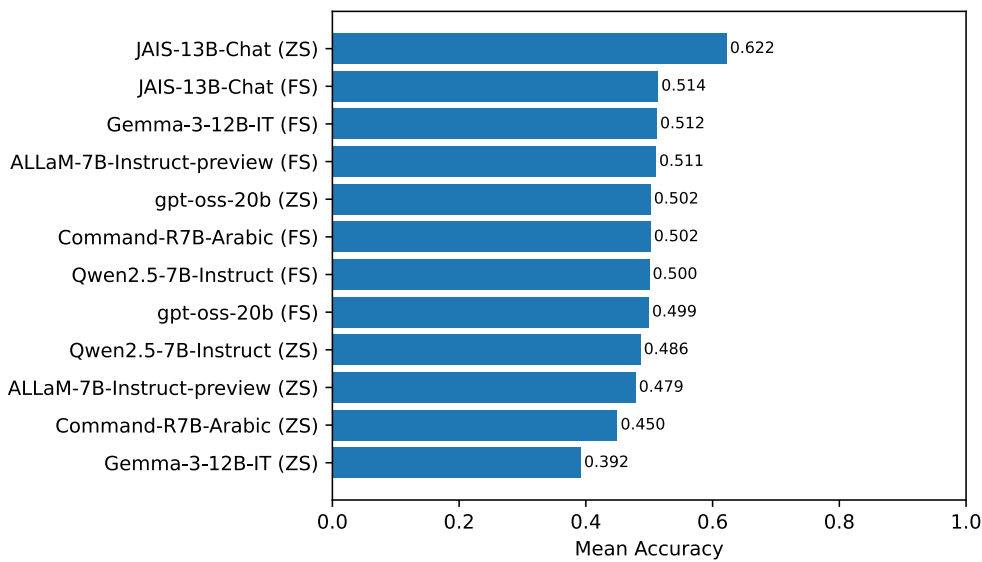

Figure 7: LLMs (ZS/FS): JAIS-13B-Chat is strongest, most models hover near chance.

The table also demonstrates that traditional classifiers provide solid baselines, with Linear-SVC (0.827) and LR (0.813) as leading models. Tree-based models such as RF and Light-GBM presented slightly less performance, however remain competitive. The results indicate that lightweight models can offer dependable accuracy with narrow CI, yet are consistently outperformed by transformer encoder models.

Mean results show that prompting LLMs underperform supervised approaches as mean evaluation results, Jais (ZS) shows the strongest performance within all prompting-based models with mean accuracy of 0.62, while most ZS/FS performance poorly with accuracies ranging between 0.39-0.51. This reflects the difficulty of classifying Arabic LLM vs Human texts in cross-genre setups. This supports our previous analysis that prompting alone is insufficient for a robust detection.

| Model | Accuracy | Macro F1 | ROC__AUC |
|---|---|---|---|
| **BERT-based (Transformers)** | | | |
| AraBERTv2-Large | $0.923 \pm 0.028$ | $0.924 \pm 0.088$ | $0.989 \pm 0.012$ |
| AraElectra | $0.917 \pm 0.112$ | $0.917 \pm 0.121$ | $0.989 \pm 0.014$ |
| AraBERTv2-Base | $0.913 \pm 0.098$ | $0.913 \pm 0.091$ | $0.990 \pm 0.016$ |
| XLM-R-Large | $0.913 \pm 0.097$ | $0.915 \pm 0.098$ | $0.990 \pm 0.009$ |
| Asafaya-BERT-Large | $0.902 \pm 0.111$ | $0.903 \pm 0.116$ | $0.987 \pm 0.014$ |
| XLM-R-Base | $0.894 \pm 0.109$ | $0.893 \pm 0.116$ | $0.980 \pm 0.031$ |
| MARBERT | $0.890 \pm 0.113$ | $0.893 \pm 0.110$ | $0.978 \pm 0.016$ |
| Asafaya-BERT-Base | $0.889 \pm 0.127$ | $0.889 \pm 0.135$ | $0.983 \pm 0.019$ |
| ARBERv2 | $0.888 \pm 0.122$ | $0.888 \pm 0.125$ | $0.976 \pm 0.039$ |
| Google-mBERT | $0.887 \pm 0.154$ | $0.889 \pm 0.103$ | $0.973 \pm 0.019$ |
| **Traditional** | | | |
| LinearSVC | $0.827 \pm 0.020$ | $0.809 \pm 0.020$ | $0.907 \pm 0.037$ |
| LogisticRegression | $0.813 \pm 0.112$ | $0.798 \pm 0.148$ | $0.916 \pm 0.067$ |
| RandomForest | $0.800 \pm 0.101$ | $0.784 \pm 0.138$ | $0.918 \pm 0.037$ |
| ComplementNB | $0.794 \pm 0.085$ | $0.781 \pm 0.112$ | $0.898 \pm 0.062$ |
| LightGBM | $0.789 \pm 0.086$ | $0.775 \pm 0.088$ | $0.895 \pm 0.057$ |
| **LLMs (ZS/FS prompting)** | | | |
| JAIS-13B-Chat (ZS) | $0.622 \pm 0.043$ | $0.620 \pm 0.044$ | $0.656 \pm 0.059$ |
| JAIS-13B-Chat (FS) | $0.513 \pm 0.032$ | $0.386 \pm 0.097$ | $0.517 \pm 0.042$ |
| Gemma-3-12B-IT (FS) | $0.512 \pm 0.039$ | $0.364 \pm 0.085$ | $0.516 \pm 0.054$ |
| ALLaM-7B-Instruct-preview (FS) | $0.510 \pm 0.031$ | $0.384 \pm 0.094$ | $0.516 \pm 0.051$ |
| gpt-oss-20b (ZS) | $0.502 \pm 0.003$ | $0.343 \pm 0.012$ | $0.489 \pm 0.086$ |
| Command-R7B-Arabic (FS) | $0.502 \pm 0.010$ | $0.344 \pm 0.046$ | $0.503 \pm 0.014$ |
| Qwen2.5-7B-Instruct (FS) | $0.500 \pm 0.001$ | $0.342 \pm 0.031$ | $0.501 \pm 0.007$ |
| gpt-oss-20b (FS) | $0.499 \pm 0.002$ | $0.334 \pm 0.002$ | $0.494 \pm 0.025$ |
| Qwen2.5-7B-Instruct (ZS) | $0.486 \pm 0.018$ | $0.373 \pm 0.039$ | $0.336 \pm 0.074$ |
| ALLaM-7B-Instruct-preview (ZS) | $0.478 \pm 0.020$ | $0.353 \pm 0.009$ | $0.436 \pm 0.044$ |
| Command-R7B-Arabic (ZS) | $0.449 \pm 0.030$ | $0.348 \pm 0.012$ | $0.353 \pm 0.025$ |
| Gemma-3-12B-IT (ZS) | $0.392 \pm 0.031$ | $0.337 \pm 0.023$ | $0.341 \pm 0.077$ |

Table 10: Overall performance across families: Accuracy, macro-F1, ROC__AUC results across SANAD, ANAD, MDAT, MASC, HARD, BRAD. Values are mean $\pm$ 95% CI.

### 9.6. Effect of dataset size

Scaling from 10% to 100% of the dataset improved the performance for both BERT and traditional models, while LLMs in ZS and FS showed no benefit. BERT mean accuracy increased from 0.892 to 0.911, and traditional models improved from 0.792 to 0.816, confirming that both model categories benefit on additional data in training and evaluation. In contrast, LLMs ZS and FS showed no improvement with 0.499 regardless of dataset size as shown in Figure 8(a)–(c).

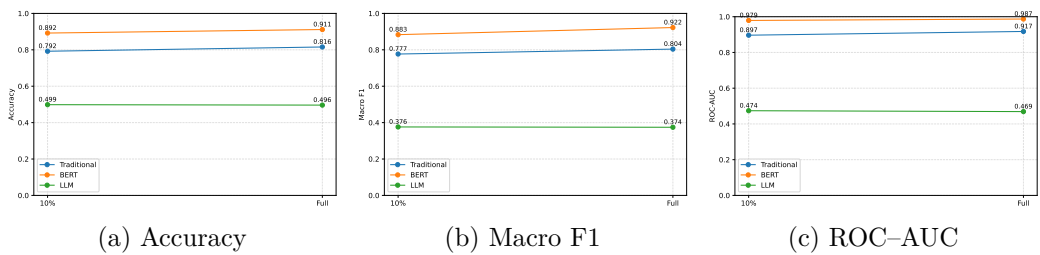

| (a) Accuracy | (b) Macro F1 | (c) ROC–AUC |
|:---:|:---:|:---:|

Figure 8: Effect of dataset size (10% vs Full) across families. Traditional and BERT improve across metrics, LLMs remain flat.

### 9.7. Genre and split robustness

Performance varied notably across dataset splits which reflects data, source, or genre shifts as shown in Figure 9, which shows the mean of accuracy, macro f1, and ROC_AUC. BERT models excelled on MASC (0.989) and "All" splits (0.973), but dropped significantly on SANAD+ANAD (0.617). Traditional models showed a similar pattern, with strong performance on "All" (0.894) and MASC (0.876) but weak performance on SANAD+ANAD (0.546). LLMs not only remained consistently poor and less interactive across splits ( 0.48-0.52). This highlights the importance of genre diversity in training data and the risks of deploying models in unseen text genres.

Our analysis reveals that robustness challenges are not only linked to the source genre (e.g. news, social media, reviews) but also to structural and linguistic factors. For example, token count distributions vary across sources (Table 6), with short tweets in MDAT/MASC and very long scripts in SANAD/ANAD, which likely contributes to shift in performance. In addition, variations across MSA and DA add further difficulty. These results suggest that cross-genre robustness is impacted by genre, sentence lengths, and dialects.

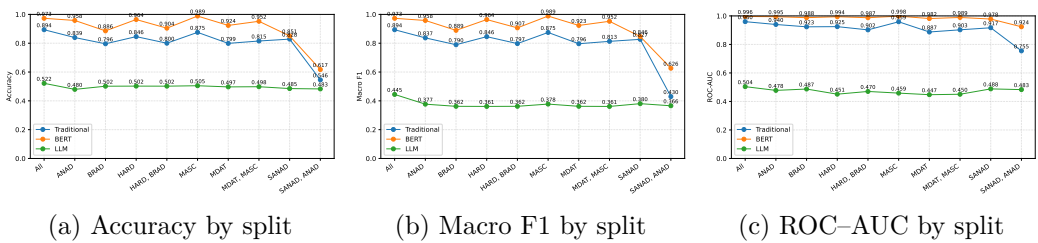

(a) Accuracy by split  (b) Macro F1 by split  (c) ROC–AUC by split

Figure 9: Genre robustness across dataset splits for each family. Performance peaks on social content (e.g., MASC) and drops on news (SANAD/ANAD), especially for LLMs.

### 9.8. Best and worst models

The best single run overall was achieved by XLM-R-Base with 0.994 accuracy, followed closely by AraElectra (0.993) and AraBERTv2 variants. Within traditional models, Logistic Regression achieved a peak of 0.930. The best LLM, JAIS-13B-Chat ZS reached 0.672. In contrast, the worst performances was obtained by Gemma-3-9B ZS with accuracies as low as 0.348. These comparisons reinforce the superiority of BERT-based models for Arabic LLM and Human classification tasks (Figure 10).

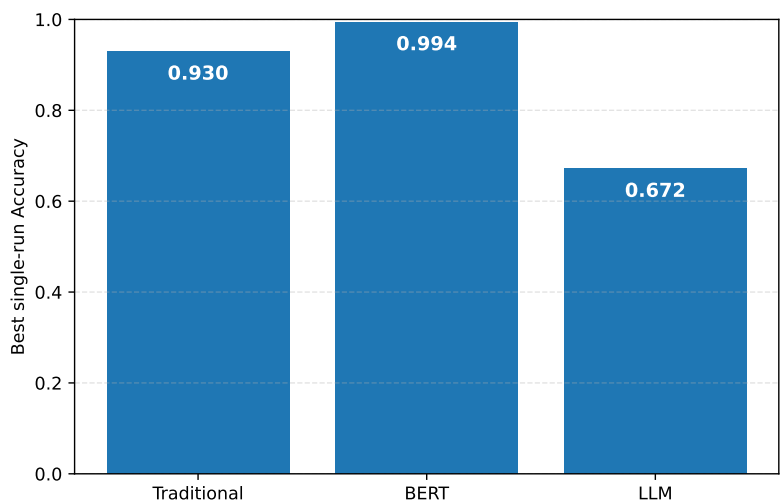

Figure 10: Best single-run accuracy per family: BERT approaches perfection, traditional is strong, LLMs substantially weaker.

### 9.9. Error analysis

Figure 11 shows mean precision and recall by family, a detailed error analysis revealed that LLMs frequently achieved poor mean precision of 0.41 across 240 LLM runs (min= 0.0, max= 1.0), resulting in frequent false predictions where human text is misclassified as LLM-generated. By comparison, transformer encoders averaged at 0.92 and traditional 0.80. In addition, an analysis of prediction errors revealed that common errors appear in similar model categories. BERT-based models scored balanced mean recall of 0.92 and precision of 0.93, making them highly dependable. Traditional models also showed high recall (0.88) but lower precision (0.80). LLMs demonstrated the opposite, precision was showing low chances (0.41), but recall collapsed to 0.37, leading to frequently undetected LLM-generated texts.

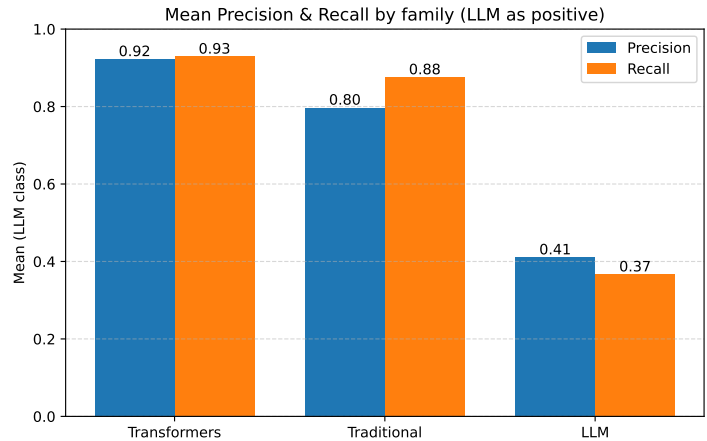

Figure 11: Mean precision by family (LLM treated as the positive class).

### 9.10. Achievements and Challenges

- BERT-based models remain the most reliable detectors when tailored for Arabic, yet their performance still drops significantly on certain unseen genres.

- ZS/FS LLM prompting is ineffective for Arabic LLM- and Human-generated text classification, highlighting limitations when relying on out-of-the-box LLMs.

- Increasing the data size improves traditional ML/BERT performance but not LLM prompting models, which reveals that data alone is not enough, and that model adaptation matters.

- All traditional and transformer models are indicating poor generalization in some sources, which highlights that true cross-genre robustness remains an open challenge.

- Errors show opposite failure modes: BERT balanced precision/recall, Traditional high precision but poor recall, LLMs poor precision. This guides where future models need to improve, and the possibility to fill gaps when interconnected through an advanced pipeline.

- When using LLM prompting, models can produce null, empty, duplicated, or confusion when generating responses. This can be overcome by setting a strict prompt and pre-processing with enforcing sanity checks before and after each step to ensure traceability and reproducibility.

## 10. Conclusion

This work introduces ALHD, a novel large-scale comprehensive Arabic dataset explicitly labelled for human- and LLM-generated texts. We constructed ALHD using six data sources across three different genres, and augmented them with outputs from three top-tier LLMs, resulting in a multi-genre corpus that fills a critical gap in Arabic NLP. Alongside the dataset, we conducted extensive benchmarking experiments, The results show that fine-tuned Arabic BERT-based models can achieve very high detection accuracy, often exceeding 90-95% and approaching perfect classification multigenre experiments. These models outperformed both traditional ML and ZS/FS LLM prompts by a wide margin, which shows the effectiveness of fine-tunning for Arabic LLM detection. However, our cross-genre evaluations demonstrated poor performance when confronted with unseen data in different genres or dialects. This highlights the necessity of having a robust and efficient detection approach.

Overall, ALHD establishes a strong foundation for trustworthy AI research in Arabic. By publicly releasing both the dataset and the benchmarking code, we enable the community to build upon our work and address existing research gaps by furthering research in Arabic LLM-generated text detection. We believe that ALHD will facilitate new studies on mitigating LLM misuse in Arabic, and we hope it will inspire the progress toward secure and reliable Arabic NLP in the era of generative AI.

The ALHD dataset is publicly available, along with preprocessing scripts, splits, and code used in benchmarking. Access to the dataset is provided through Zenodo at `https://doi.org/10.5281/zenodo.17249602` [57], while the benchmarking experiments codebase is hosted on GitHub at `https://github.com/alikhairallah/ALHD-Benchmarking`. Researchers can reproduce all 540 experiments via provided configurations and logging utilities. The dataset is intended to be used for academic and non-commercial use, with restrictions to prevent misuse in adversarial text generation.

## 11. Limitations

While ALHD and the associated experimentation we provide in this study bring new resources and insights for furthering research in Arabic LLM-generated text detection with a primary focus on model generalization, this is not without some limitations.

***Dataset limitations..*** There are some limitations linked to the dataset characteristics and its coverage:

- Although ALHD is multi-genre, the fact that it is limited to three genres means that some other genres such as academic essays or books are not covered; the dataset still serves its purpose of enabling cross-genre experimentation toward evaluating generalizability.

- The dataset does not cover all existing Arabic dialects (focus on MSA + limited DA). Some dialects might not be present.

- While we aimed to cover a diverse set of LLM generators, this is limited to three, with potential limitations with respect to dealing with other models or newer models.

***Experimental limitations..*** There are some limitations with the experimentation and analysis conducted in this study, which can be further expanded in future work:

- LLM benchmarks are limited to prompting, and may exhibit better results after fine-tuning or model adaptation, which is beyond the scope of our study.

- Resources and time constraints prevented testing very large state-of-the-art models, yet our focus here has been on providing sufficient benchmarking results and analysis.

- Models are sensitive not just to genres, but also to text length and dialect, which means there is still room for improvement in terms of robustness.

## 12. Future Work

Building on the presented dataset and benchmarking results, several improvements remain critical for advancing Arabic LLM detection. We outline the following future work directions:

***Dataset expansion..*** While ALHD covers multiple genres and dialects, extending it further would enhance robustness. Future iterations could incorporate additional field like adversarial annotation where an adversarial texts would be added to assess model in detecting sophisticated social engineering, phishing, and smishing attacks.

***Detection approaches..*** The benchmarking highlighted the limitations of both classical and prompting LLM-based models. Future research could explore more robust approaches for human- and LLM-generated texts, and compare them against best evaluation performance in the benchmarking performed.

***Robustness testing.*** Future evaluations benchmarking should apply on cross-LLM generalization (detecting outputs from unseen models) and cross-dialect evaluation, this would better reflect real-world challenges where detectors face diverse and shifting distributions of text.

***Deployment considerations..*** To support practical adoption in cybersecurity, education, and digital forensics, future work should focus on exploring lightweight and efficient detectors which are suitable for real-time or resource-constrained environments.

## 13. Ethical Considerations

The construction of ALHD followed a strict ethical and standards to ensure both privacy preservation and responsible research use. All human-written texts were sourced from publicly available Arabic dataset (SANAD, ANAD, MDAT, MASC, HARD, BRAD), which are already widely used in NLP research. No private or personally identifiable information was collected including user-specific metadata. While ALHD is designed to support the development of trustworthy AI and enhance defence against cyberthreats such as misinformation, phishing, and academic dishonesty, it also carries the potential for misuse in training adversarial models that generate humanized outputs. To mitigate this, ALHD is released under the Creative Commons Attribution-NonCommercial 4.0 (CC BY-NC 4.0) license with explicit restrictions against commercial or malicious applications, while experiments codebase is released under the MIT license. In addition, all LLM-generated texts were produced through official APIs (OpenAI GPT-3.5-turbo, Google Gemini 2.5, Cohere Command-R) using default parameters, ensuring realistic generation behaviour that reflects typical user interaction rather than adversarial settings. By adhering to these principles, ALHD seeks to maximize responsible research practices in Arabic NLP research and applications.

## Acknowledgment

This research utilized Queen Mary's Apocrita HPC facility, supported by QMUL Research-IT, we also acknowledge and appreciate the assistance of the ITS Research team at Queen Mary University of London [50].